\documentclass{article}

\usepackage{arxiv}

\usepackage[utf8]{inputenc} 
\usepackage[T1]{fontenc}    
\usepackage{hyperref}       
\usepackage{url}            
\usepackage{booktabs}       
\usepackage{amsfonts}       
\usepackage{nicefrac}       
\usepackage{microtype}      
\usepackage{lipsum}
\usepackage{graphicx}
\graphicspath{ {./images/} }
\usepackage[T1]{fontenc}
\usepackage{color}
\usepackage{enumitem}
\usepackage{caption}
\usepackage{subcaption}
\usepackage{multicol}
\usepackage{hyperref}
\usepackage{cite}  
\usepackage{orcidlink}

\title{Surrealistic-like Image Generation with Vision-Language Models}

\author{
Elif Ayten\orcidlink{0009-0005-4954-0475} \\
 Vrije Universiteit Amsterdam, \\
 1081 HV Amsterdam, the Netherlands \\
 \texttt{e.ayten@student.vu.nl} \\
   \And
 Shuai Wang\orcidlink{0000-0002-1261-9930} \\
   Vrije Universiteit Amsterdam, \\
 1081 HV Amsterdam, the Netherlands \\
  \texttt{shuai.wang@vu.nl} \\
  \And
 Hjalmar Snoep\orcidlink{0000-0002-6072-4073} \\
 Snoep Studio,\\
 7322 DS Apeldoorn, the Netherlands \\
  \texttt{snoepstudio@gmail.com} \\
}

\begin{document}
\maketitle
\begin{abstract}
Recent advances in generative AI make it convenient to create different types of content, including text, images, and code. In this paper, we explore the generation of images in the style of paintings in the surrealism movement using vision-language generative models, including DALL-E, Deep Dream Generator, and DreamStudio. Our investigation starts with the generation of images under various image generation settings and different models. The primary objective is to identify the most suitable model and settings for producing such images. Additionally, we aim to understand the impact of using edited base images on the generated resulting images. Through these experiments, we evaluate the performance of selected models and gain valuable insights into their capabilities in generating such images. Our analysis shows that Dall-E 2 performs the best when using the generated prompt by ChatGPT. 
\end{abstract}


\section{Introduction}
\label{sec:intro}

 The ability of Artificial Intelligence (AI)  to generate images from text has been an active field of research over the past few years. For example, OpenAI's DALL·E \cite{dalle} is a generative model that produces high-quality images from textual descriptions (i.e. prompts).  As technology develops, more and more tools are available.  DreamStudio \cite{DreamStudio}, Deep Dream  Generator \cite{deepdreamgenerator}, MidJourney \cite{midjourney}, and various other models are getting more and more powerful in this domain. These models typically require users to input text, in the form of words or longer sentences, to generate images. Alternatively, some models can also take a base image along with the text input to facilitate the generation process. Such progress shows a tendency towards realism, rather than enhancing flexibility and novelty in artistic expression. Some tools make it possible to create some diverse and realistic visuals (e.g. using Generative adversarial networks (GANs) \cite{karras2017progressive}), but at times, the generated objects in these images can exhibit a lack of logical coherence, both in their selection and how they are combined or the style of assemblage. However, this feature can serve as a valuable source of inspiration for artists \cite{esman2011psychoanalysis, Gurney_2022}.

Surrealism was an artistic movement that gained popularity between WWI and WWII. It emerged as an influential artistic movement following advances in modern psychology, especially influenced by the work on unconscious by Sigmund Freud \cite{esman2011psychoanalysis}. The Surrealists sought to harness the creativity of the unconscious mind. Some of the previous studies claim that these generative models can be considered getting better at generating art that emulates surrealism and fantasy \cite{Gurney_2022}. However, this claim was not justified with experimental evaluation. The generated images can exhibit creative potential in unconventional arrangements by being used as a basis for supernatural imagery or potential ideas to have objects \textit{juxtaposed} (placing different things side by side for contrasting effect) \cite{Lanz_2023}.  Our cursory examination shows that some of these generated images can stand alone as thought-provoking pieces, showcasing their artistic value. We aim to explore how generated images can serve as thought-provoking creations, forging connections with the depths of our unconscious mind.

\begin{figure}[!htb]
  \centering
  \begin{subfigure}[t]{0.18\textwidth}
    \centering
    \includegraphics[width=\linewidth]{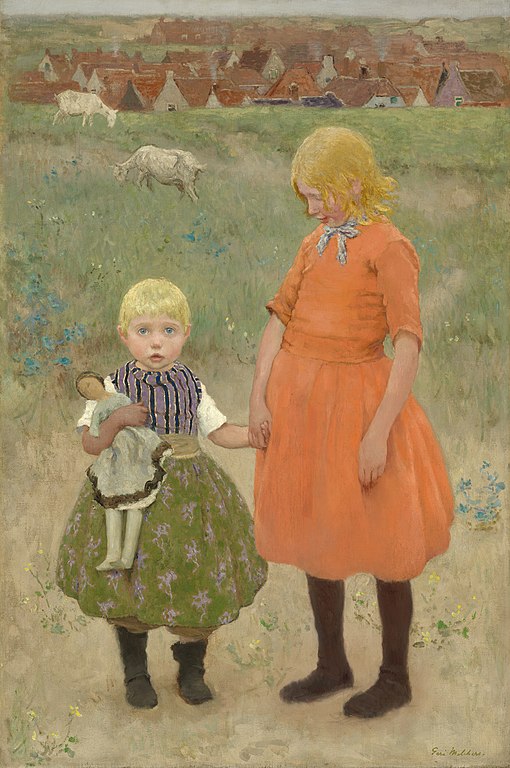}
    \caption{The painting by Gari Melchers (The Sisters, 1895)}\label{fig:originalImageFig1}
  \end{subfigure}\hfill
  \begin{subfigure}[t]{0.18\textwidth}
    \centering
    \includegraphics[width=\linewidth]{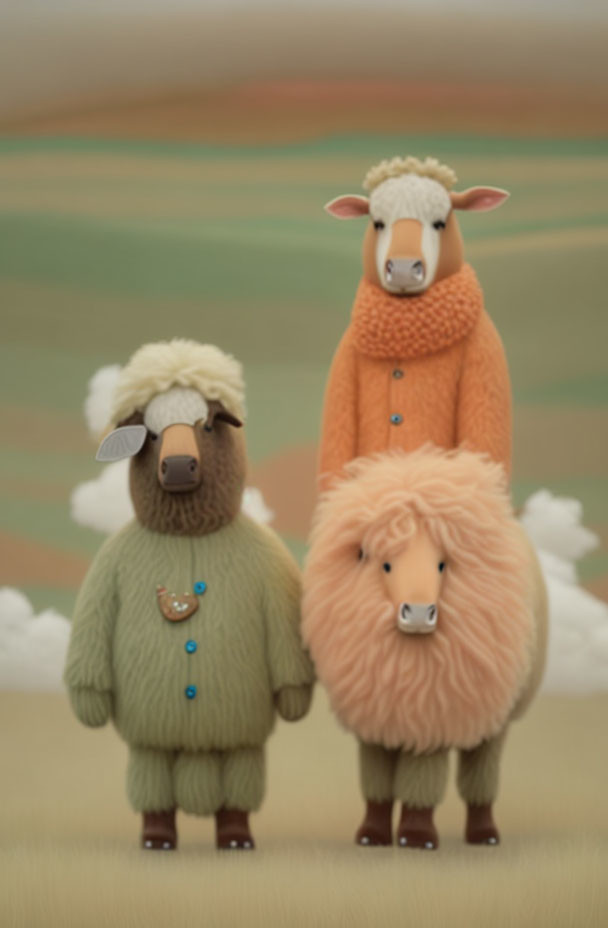}
    \caption{An image generated by  Deep Dream Generator}\label{fig:deepdreamfig1}
  \end{subfigure}\hfill
  \begin{subfigure}[t]{0.20\textwidth}
    \centering
    \includegraphics[width=\linewidth]{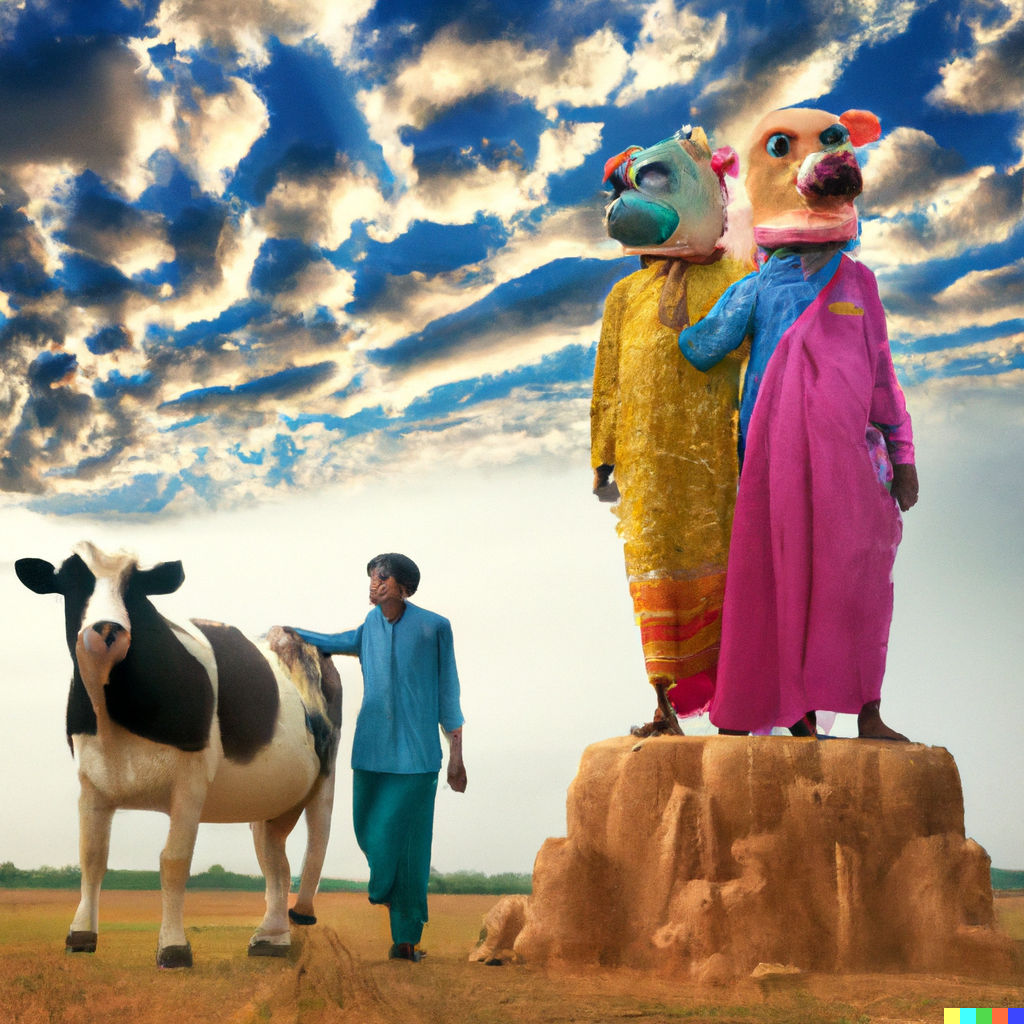}
    \caption{An image generated by  DALL·E 2}\label{fig:dallefig1}
  \end{subfigure}\hfill
  \begin{subfigure}[t]{0.20\textwidth}
    \centering
    \includegraphics[width=\linewidth]{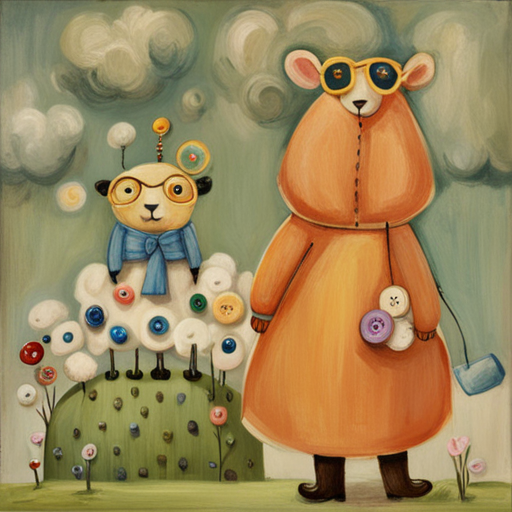}
    \caption{An image Generated by  DreamStudio}
    \label{fig:DreamStudiofig1}
  \end{subfigure}
  \caption{The original painting and some generated images in diverse styles of image generation using different models and corresponding text. Each model's distinct approach yields varied styles of output for image captions.
}\label{fig:Generated4images1figure}
\end{figure}

The main objective of this research paper is to study the generation of images by using vision-language models. We focus on the text-to-image generation and the transformation of realistic images to images in the style of paintings in the surrealism movement. The term `surrealistic painting' is often used to refer to paintings in the Surrealism movement. Given the lack of a well-established definition, in this paper, we let the term \textit{surrealistic-like} refer to objects that exhibit the following features. These features were extracted from a survey of literature and the official websites of well-known museums and specified galleries.

\begin{itemize}
    \item \textbf{dreamlike atmosphere}: Art pieces that seek to evoke a dreamlike or hallucinatory quality, blurring the boundaries between reality and imagination \cite{SurrealismTate}.
    \item \textbf{unexpected juxtapositions}: Art pieces that feature unexpected combinations of unrelated objects, images, or ideas, creating a sense of surprise and mystery \cite{momaMoMASurrealist}.
    \item \textbf{symbolism and metaphor}: Art pieces that employ rich symbolism and metaphor to convey hidden meanings and explore psychological and philosophical concepts \cite{metmuseumSurrealismEssay}. 
    \item \textbf{subversion of logic}: Art pieces that challenge conventional logic and rationality, embracing absurdity and irrationality as a means of expressing the subconscious \cite{thoughtcoTheseArtists}.
    \item \textbf{unconscious exploration }: Art pieces that explore the realm of the unconscious mind, tapping into dreams, fantasies, and irrational thoughts \cite{guggenheimGUGGENHEIMMUSEUM}.
\end{itemize}

Our approach is to generate surrealistic-like images from realistic base images. Take the paintings and images in Figure \ref{fig:Generated4images1figure} for example. Figure \ref{fig:originalImageFig1} is the selected base image: the original painting named The Sisters by Gari Melchers in 1895. To its right are images generated by different means. Our research aims to determine the optimal settings for generating surrealistic-like paintings. Since there are currently no established tools to definitively judge the quality of surrealistic-like images, we conducted a survey to gather opinions from artists. This approach allows us to evaluate and assess which settings produce the best surrealistic-like artwork. Thus, our main  research questions of the paper are the following: 

\textbf{RQ: How can we create surrealistic-like paintings using generative vision-language models?}

We take two approaches. The first approach takes texts and the selected base images as input. For comparison, we take only text as input. Our research question is focused on assessing how to optimize the use of existing image generation models effectively. To address our primary research question, we utilize DALL·E2, DreamStudio, and Deep Dream Generator, as they provide capabilities for text-to-image and text-image-to-image generations. We try to identify the most suitable approach and model for creating surrealistic-like paintings. Additionally, we explore the potential of large language models (e.g. ChatGPT) to enhance the performance of image-generative tools automatically. We answer this research question by studying the following sub-research questions (SRQ):

\textbf{SRQ1: Which image generative approach is best suited for the generation of surrealistic-like images?}


\textbf{SRQ2: Can generative language models make meaningful prompts for the generation of surrealistic-like images?}


\textbf{SRQ3: How can we improve the generation of surrealistic-like images?}

To address our research questions, we have an approach that involves conducting three experiments with a survey to evaluate the obtained results. The main contribution of this paper is as follows. 
\begin{enumerate}
\item A dataset\footnote{The dataset, the code, and the survey and its results are in the repository as open source \url{https://github.com/ElifAyten/ElifAytenThesis2023}.} of 235 images was generated using three image generators under different settings. In addition, we generated 12 downscaled images and 12 burred images during our experiments. Finally, we generated 16 images as a showcase for future work.

\item An evaluation of the generated images to determine their suitability for creating surrealistic-like artwork. 

\item The paper also presents some primitive steps towards optimizing the settings for the generation of compelling surrealistic-like images.
\end{enumerate}

This paper is organized as follows:
Section \ref{sec:related} presents related work and introduces state-of-the-art image generation techniques. Section 
\ref{sec:method} presents the methodology of this paper, including the models, base images, setup, survey, and image generation. Following that, Section 
\ref{sec:evaluation} analyzes the evaluation results. 
Section \ref{sec:Discussion} encompasses an extensive analysis of the results and the discussion.
Finally, the conclusion and future work are included in Section \ref{sec:Conclusion}.

\section{Related Work}
\label{sec:related}
\subsection{Artistic Image Generation }
Image generation has witnessed some remarkable progress due to the availability of a large number of datasets and the advances of various large-scale generative AI methods. Among the leading tools in this field are DALL-E \cite{dalle}, Midjourney \cite{midjourney}, and Stable Diffusion \cite{altexsoftImageGeneration}. DALL-E has a simple interface and it is a text-to-image model \cite{Boehman_2023}. In comparison, Midjourney is more complex and requires its users to use a specified platform (without an API). The stable diffusion model which we use is DreamStudio and can be hit or a miss with the generation depending on the version \cite{Boehman_2023}. These tools use various generative AI techniques, including Generative Adversarial Networks (GANs) \cite{karras2017progressive}, Transformers \cite{vaswani2017attention}, and Variational Autoencoders (VAEs) \cite{kingma2022autoencoding} for the generation of images.

As for evaluation, the Inception Score and Fréchet Inception Distance (FID) are algorithms used to assess the performance of generative image models. The FID score evaluates the quality and diversity of generated images by comparing their distribution with real images. It quantifies the dissimilarity between the feature representations of these two distributions \cite{kim2020simplified}. The Inception Score, conversely, focuses on assessing the recognizability of objects within the generated images \cite{barratt2018note}. Both algorithms provide valuable insights into the capabilities of generative image models.

Another approach involves human evaluation, incorporating data and results that rely on human perception and understanding. However, it is important to note that the methods used for human evaluation metrics can vary among researchers. The qualifications of assessors may not be specified, or the survey and evaluation processes themselves might be overly complex or overly simplistic \cite{otani2023toward}. This can lead to differences in the results.

\section{Methodology}
\label{sec:method}
This section presents an overview of the models utilized in this study (Section \ref{sec:models}) and the base images for the models (Section \ref{sec:baseimages}), the setup for the prompt generations(Section \ref{sec:setup}). Additionally, an explanation of the image generation processes for each part of the survey (Section \ref{sec:image}). Finally, the survey design in Section \ref{sec:survey}.

\subsection{Models}
\label{sec:models}
In our study, we utilize three models for image generation. DALL-E is limited to generating images solely from text input, while Deep Dream Generator and DreamStudio have the capability to generate images from both text and base image inputs. The \textbf{Deep Dream Generator} is a generative AI solution focused on image generation \cite{deepdreamgenerator}. Deep Dream Generator utilizes deep learning techniques to generate visually appealing and imaginative images based on given inputs \cite{deepdreamgenerator}. \textbf{DALL·E}, a deep learning model by OpenAI, is designed for generating images from textual prompts \cite{dalle}. It uses a text-image pair dataset and is built on the GPT (Generative Pre-trained Transformer) architecture \cite{dalle}. \textbf{DreamStudio} is an interface that utilizes the Stable Diffusion Model, a text-to-image model capable of generating images based on textual inputs \cite{DreamStudio}. The Stable Diffusion Model incorporates a forward and reverses diffusion process, where an image is transformed into a noisy version in the forward process and then restore the image to its pre-noise state in the reverse process \cite{DreamStudio}. 



\subsection{Base Images}
\label{sec:baseimages}

Our input images consist of a selection of six base images that are curated from renowned realist artists, including Gustave Courbet, Rosa Bonheur, and Meredith Frampton. We deliberately include various realistic styles such as animal realism, satirical realism, and portrait realism. Table \ref{tab:base} provides details about the painting names, the artists' styles, and the respective years of creation.

\begin{table}[h]
\centering
\footnotesize
\begin{tabular}{lp{3cm}ll}
\textbf{Painting} & \textbf{Painter} & \textbf{Style} & \textbf{Year} \\ \hline
The Sisters  & Gari Melchers & Portrait Realism & 1895 \\
 Greyhounds of Comte de Choiseul   &Gustave Courbet  & Animal Realism &1866 \\
The Horse Fair & Rosa Bohnear & Animal Realism & 1853 \\
Horse in the Sunlight& Guiseppe Abbati & Animal Realism & 1866 \\
The Aristocrat's Breakfast & Pavel Fedotov& Satirical Realism & 1850 \\
 Sir Clive Forster-Cooper& George Vernon Meredith Frampton  & Portrait Realism & 1945
\end{tabular}
\caption{Comparison of the base paintings: name, artist, style, and year of creation.}
\label{tab:base}
\end{table}

\subsection{Experimental Setup}
\label{sec:setup}
For our experiments, we use input prompts consisting of realistic base images. To label the objects within these images, we employ the YOLO algorithm \cite{v7labs:YoloObjectDetection}, which is an object detection tool. YOLO is known for its speed and accuracy.

\begin{figure}[ht!]
    \centering
    \subcaptionbox{%
        ``The Aristocrat's Breakfast'' with YOLO labeled objects and their probabilities: 1 person (Probability: 0.95), 1 dog (0.93), 2 chairs (0.80 and 0.92), 1 couch (0.50), 1 dining table (0.55), 1 book (0.30), 1 vase (0.89).%
        \label{subfig:aristocrat}
    }{%
        \includegraphics[width=0.44\textwidth]{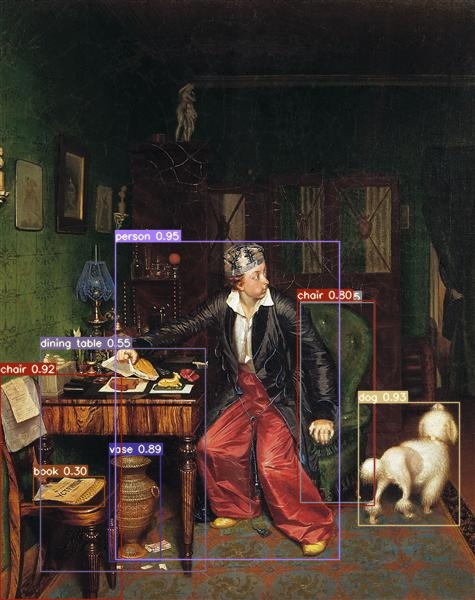}%
    }\hfill
    \subcaptionbox{%
        ``Sisters'' with its detected objects and their probabilities using YOLO: 1 person (0.95), 1 person (0.96), 1 teddy bear (0.83), 1 sheep (0.42), 1 sheep (0.59).%
        \label{subfig:sisters}
    }{%
        \includegraphics[width=0.40\textwidth]{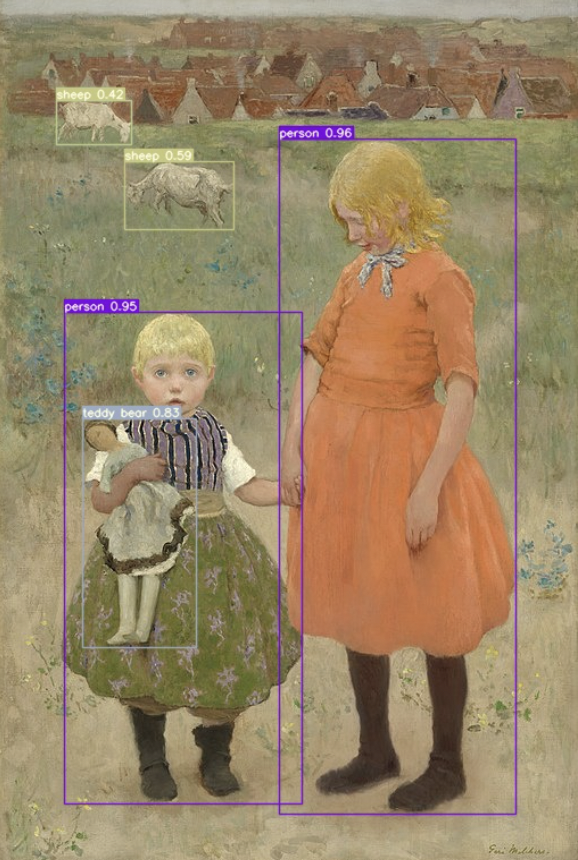}%
    }
    \caption{YOLO labeled images}
    \label{fig:awesome_image2}
\end{figure}

Once the labeling process is completed, we use ChatGPT-3, to create longer and more detailed prompts based on the extracted object labels. For example, for Figure  2, we have the input as ``Generate a prompt for surrealistic image generation within 15 words that contains the following objects: 1 teddy bear,2 sheep and 2 person''. The generated descriptions are as follows. \emph{Prompts of 15 words}\footnote{All the prompts in this paper were generated on June 3rd, 2023.} each: Two persons encounter two sheep, a cow, and a teddy bear amidst a floating garden
\emph{Prompts of 50 words} each:Two persons, one adorned with a cow mask, guide two sheep, each wearing oversized spectacles, towards a towering teddy bear monument. The sky weaves patterns of wool, while fluffy clouds rain streams of multicolored buttons, creating a surreal landscape of whimsical companionship.

\subsection{Image Generation}
\label{sec:image}

We conduct our image generation study in three settings described in the three following parts.

\subsubsection{Part 1: Comparative analysis of multiple generative models}
\label{sec:Part 1}
The initial part of our study aims to compare three image generation models: Deep Dream Generator, DreamStudio, and DALL-E (version 2). We use different inputs for each model. For DeepDream and DreamStudio, we try both base images and diverse prompts. As for DALL-E, we focus on various prompt settings due to its unique nature. Our goal is to evaluate their abilities in producing surrealistic images by adjusting prompt types and lengths. In the first part, we employ three key setups for image generation. First, we create images using YOLO labels as prompts—simple words describing objects in the images. Second, we compare the models using 15-word prompts generated by ChatGPT, a proficient language model. Finally, we explore ChatGPT prompts containing 15 words and the artist's name from the base image. This extra prompt component adds contextual information about the artist, potentially impacting the surrealistic qualities of the images.

\subsubsection{Part 2: Comparing various prompt settings within each generative model}
We focus on testing each model individually, examining their performance with different prompt settings. We specifically explore the capabilities of the three models using the same base image under various prompt settings. The objective of the second part is to assess how each model handled different prompt lengths and determine which model performed better under specific settings. By varying the prompt lengths and analyzing the generated results, we aim to gain insights into the strengths and limitations of each model and identify the most effective prompt settings for surrealistic-like image generation. Different from the previous phase, we also include prompts of 50 words, enabling more details.  

\subsubsection{Part 3: Base image editing comparison for Deep Dream Generator and DreamStudio}
Finally, we compare the performance of the models when using edited base images versus non-edited base images. First, we apply a blurring effect to each base image, creating a set of blurred images. These blurred images, along with a 50-word long prompt, are used as input for both the Deep Dream Generator and DreamStudio models. Second, we downsize the base images to one-fourth of their original size, generating a set of downscaled images. These downscaled images, accompanied by a 50-word prompt, are compared to the results obtained using only a 50-word prompt without downsizing.

\subsection{Survey}
\label{sec:survey}

Given the absence of neural networks capable of accurately determining features in this domain, we developed a large survey that relies on subjective opinions by artists. We had 18 participants after reaching out to art students and recent graduates of some art schools and universities, including the Gerrit Rietveld Academie, Vrije Universiteit Amsterdam, Hogeschool voor de Kunsten Utrecht, and the Eugeniusz Geppert Academy of Art and Design in Wroclaw. Since the personal information of the participants was anonymized, the authors did not track the exact affiliation of each.  The questionnaire is divided into three parts, each corresponding to the image generation as described in the previous subsection.

In Part 1 of the survey, we conduct a comparative analysis of multiple generative models and settings. Our goal is to gain insights into how different settings such as YOLO labeling and prompts generated by ChatGPT with lengths of 15 words compare with 15 words with artist's name prompts. In Part 2 of the survey, our focus is on comparing different prompt settings within each generative model. We examine three models and evaluate their performance under distinct prompt configurations, including YOLO labeling and prompts generated by ChatGPT, with lengths of 15 words and 50 words and simply the YOLO labeling. In Part 3 of the survey, we introduce base image downscaling and blurring. Participants are presented with variations of the base input image for Deep
Dream generator and DreamStudio models, including downscaled versions and blurred images.

The survey comprises a total of 60 questions, with three sub-questions each: 
 
\begin{enumerate}[label=(\alph*)]
    \item  Which image(s) translates the subject matter in the original painting to something unexpected (object or element of surprise, etc.)? [multiple answers can be selected]
    
    \item  Which generated image(s) has unexpected juxtapositions of objects? [multiple answers can be selected]

    \item Which one is the best surrealistic-like painting among the three paintings, in your opinion? [only one answer  can be selected]

    \end{enumerate}



\section{Evaluation}
\label{sec:evaluation}
To ease the description of the analysis of the survey data, we introduce the following notation of prompt setting (PS) and (PIS) prompt-image setting: \textbf{PS1:} The YOLO algorithm provides information about the objects and their quantities, such as `2 dogs, 1 person', which is used in the image generation process. \textbf{PS2:} Given the PS1 and a description request, ChatGPT generates prompts with a word limit of 15. For instance, the input "The description of a surrealistic-like painting with 1 person, 2 dogs in 15 words" is used to generate images. \textbf{PS3:} Given the PS1 and a 15-word request for a description, along with the name of an artist, ChatGPT generates a prompt for image generation. For instance, the input "The description of a  surrealistic-like painting with 1 person, 2 dogs in 15 words, Gari Melchers" is used to generate images. \textbf{PS4:} Similar to PS2 but with a 50-word request for a description. For instance, the input "The description of a  surrealistic-like painting with 1 person, 2 dogs in 50 words" is used to generate images. \textbf{PIS5:} Similar to PS4 but with the base image also as input. Additionally, the base image will be downscaled to its 1/4 size. \textbf{PIS6:} Similar to PS4 but with the base image also as input. Additionally, the base image will be blurred (Gaussian blur) by using the Python Imaging Library (PIL).

\begin{table}[!t]
\centering
\label{tab:my-table}
\footnotesize
\begin{tabular}{lrrr}
Question& DALL-E 2&DreamStudio& Deep Dream Generator\\ \hline
\textbf{P1-SQA}& \underline{50.8\%} & 9.0\%  & 40.1\% \\
\textbf{P1-SQB}&\underline{50.8\%} & 8.8\%  & 40.4\% \\
\textbf{P1-SQC}&35.6\% & 12.5\% & \underline{51.9}\% \\ 
\textbf{P2-SQA}&\underline{34.6\%} & 30.7\% & \underline{34.6\%} \\
\textbf{P2-SQB}&34.2\% & 31.0\% & \underline{34.8}\% \\
\textbf{P2-SQC}&33.2\% & 32.7\% & \underline{34.2}\% \\
\textbf{P3-SQA}&N|A      & 47.3\% & \underline{52.7\%}   \\
\textbf{P3-SQB}&N|A      & 47.0\%   &\underline{53.0\%} \\
\textbf{P3-SQC}&N|A      & 49.6\% &\underline{50.4\%}
\end{tabular}
\caption{Participants' preferences for sub-question selection in each survey section, expressed as percentages, with the most chosen generator highlighted.}
\end{table}

Table 2 presents various sections of the survey, denoted as P1, P2, and P3, along with distinct sub-questions labeled SQA, SQB, and SQC. During our analysis of the responses to these sub-questions, we segregate the survey parts due to their specific content. In Part 1 and 2, all three models are featured, whereas Part 3 is exclusively about DreamStudio and Deep Dream Generator. It is evident that, concerning DALL-E, sub-questions SQA and SQB are the most frequently selected, while for DreamStudio, sub-question SQC is favored. Similarly, for Deep Dream Generator, sub-question SQC emerges as the predominant choice among participants.

\begin{figure}[ht]
    \centering
    \begin{subfigure}{0.33 \textwidth}  
        \includegraphics[width=\linewidth]{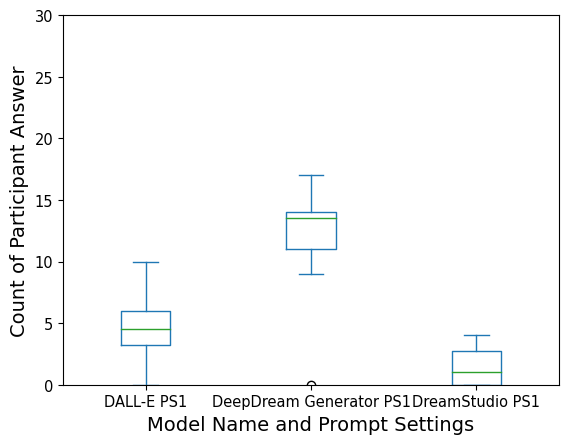}
        \caption{Participants' choices of three models regarding PS1}
        \label{fig:plot11}
    \end{subfigure}
    \begin{subfigure}{0.33\textwidth}
        \includegraphics[width=\linewidth]{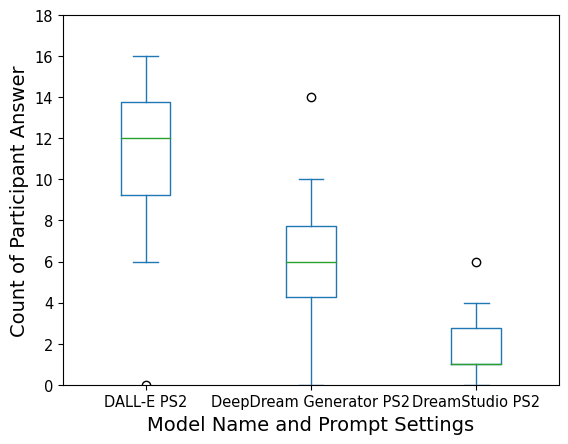}
        \caption{Participants' choices of three models regarding PS2}
        \label{fig:plot2}
    \end{subfigure}
    \begin{subfigure}{0.33\textwidth}
        \includegraphics[width=\linewidth]{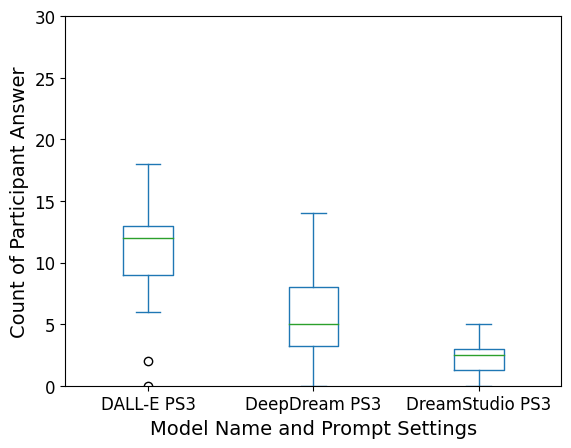}
        \caption{Participants' choices of three models regarding PS3}
        \label{fig:plot3}
    \end{subfigure}
    \caption{Survey Part 1 Results}
    \label{fig:three_plots}
\end{figure}

 Figure \ref{fig:plot11} shows the distribution of participants' answers for all the model settings. For the DALL-E PS1, the median count of participants' answers is 4.50. In contrast, the Deep Dream Generator PS1 shows a higher median count of 13.50. For the DreamStudio PS1, the median count is 1.00, indicating that it receives the least preferred overall. It is apparent that  Deep Dream Generator is the model that is selected most frequently among the participants.  The analysis of Figure \ref{fig:plot2} reveals that the DALL-E model with a 15-word prompt generated by ChatGPT was the most favored choice among the participants. Similarly, from Figure \ref{fig:plot3}, DALL-E model outperforms the other two models with a 15-word prompt generated by ChatGPT and artist name setting. Little difference was observed between PS2 and PS3 for Dall E. In summary, the Deep Dream Generator is the most preferred model for YOLO labeling while DALL-E is the most preferred tool if given textual input by ChatGPT (PS2 and PS3).

\begin{figure}[]
    \centering
    \begin{subfigure}{0.38\textwidth}
        \includegraphics[width=\linewidth]{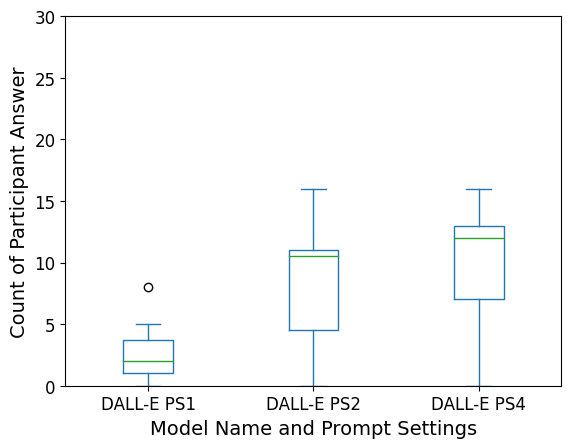}
        \caption{\footnotesize A comparison between participants' choices of DALL-E regarding PS1, PS2, and PS4}
        \label{fig:plot1.12}
    \end{subfigure}
    \hfill
    \begin{subfigure}{0.38\textwidth}
        \includegraphics[width=\linewidth]{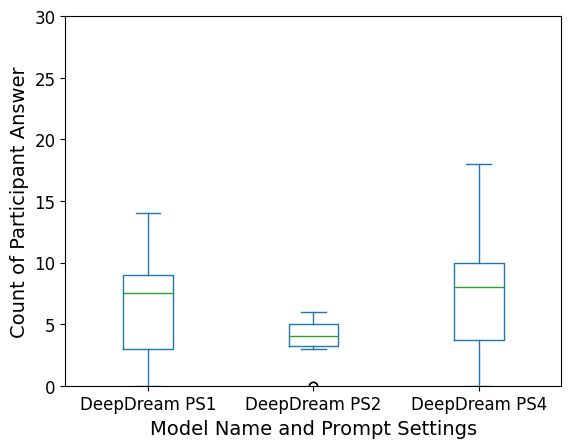}
        \caption{\footnotesize  A comparison between participants' choices of Deep Dream regarding PS1, PS2, and PS4}
        \label{fig:plot2.1}
    \end{subfigure}
    
    \vspace{0.5em} 
    
    \begin{subfigure}{0.38\textwidth}
        \includegraphics[width=\linewidth]{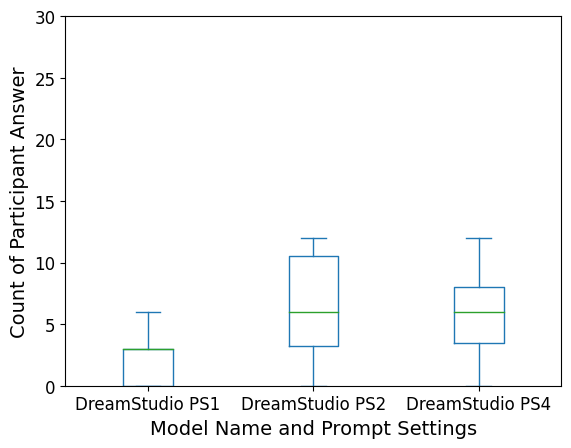}
        \caption{\footnotesize  A comparison between participants' choices of DreamStudio regarding PS1, PS2, and PS4}
        \label{fig:plot3.1}
    \end{subfigure}
    \hfill
    \begin{subfigure}{0.38\textwidth}
        \includegraphics[width=\linewidth]{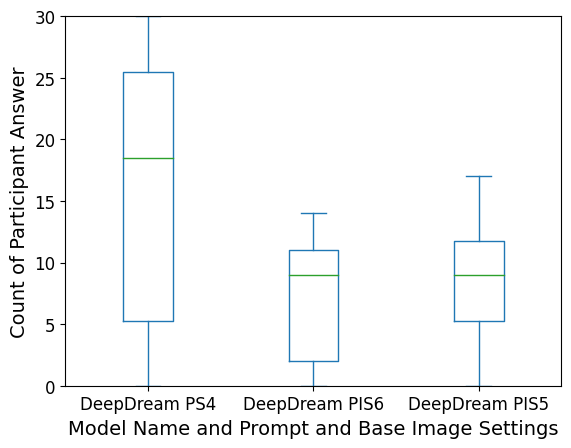}
        \caption{\footnotesize A comparison between participants' choices of Deep Dream Generator regarding PS4, PIS5, and PIS6}
        \label{fig:plot4.1}
    \end{subfigure}
    
    \vspace{0.5em} 
    
    \begin{subfigure}{0.40\textwidth}
        \includegraphics[width=\linewidth]{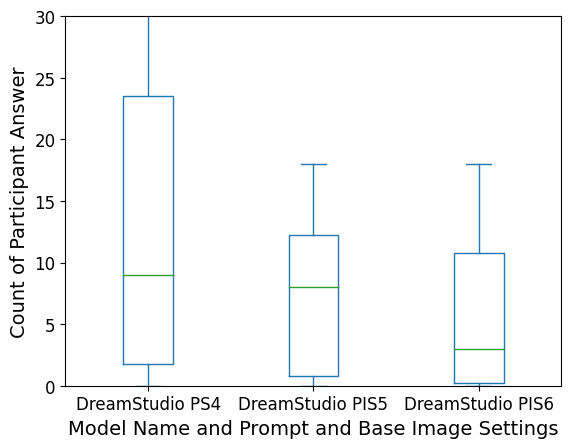}
        \caption{\footnotesize A comparison between participants' choices of DreamStudio Generator regarding PS4, PIS5, and PIS6}
        \label{fig:plot5.1}
    \end{subfigure}
    \caption{Survey Part 2 and 3 Results}
    \label{fig:five_plots}
\end{figure}

In Figure \ref{fig:five_plots}, we can observe the performance of different models under various settings. Figure \ref{fig:plot1.12} reveals that DALL-E achieves its highest median performance 12.0 with a prompt setting PS4, indicating its superior results compared to other prompt settings. The first quartile (Q1) and third quartile (Q3) values further support this effectiveness. Comparatively, the DALL-E PS2  has a median count of 10.5 while the YOLO labeling model receives the lowest median count 2.0. To summarize, DALL-E performs best with a prompt setting of 50 words.

In Figure \ref{fig:plot2.1} and Figure \ref{fig:plot3.1}, we observe that the Deep Dream Generator and DreamStudio exhibit different trends in comparison with Dall E. However, neither of them is comparable with  Dall E.  In Figure \ref{fig:plot4.1} and \ref{fig:plot5.1}, the effects of downscaling and blurring on the base image's impact on the models' performance are found to be limited. The outcomes indicate no significant enhancement compared to the original images. Notably, the downscaled iteration displays slightly superior performance compared to the blurred version, as indicated by our statistical analysis. For the Deep Dream Generator model, our analysis indicates that all three configurations exhibit similar distributions of participants' responses  \ref{fig:plot4.1} . The median scores are 18.5 for Deep Dream Generator PS4, 9.0 for Deep Dream Generator Blurred, and 9.0 for Deep Dream Generator Downscaled. In conclusion, while downscaling and blurring the base images do not lead to a significant improvement in the models' performance, the downscaled version appears to have a slight advantage over the blurred version.



\subsection{Aesthetics Analysis } 
\begin{figure}[]
  \centering
  \begin{subfigure}[!ht]{0.29\textwidth}
    \centering
    \includegraphics[width=\linewidth]{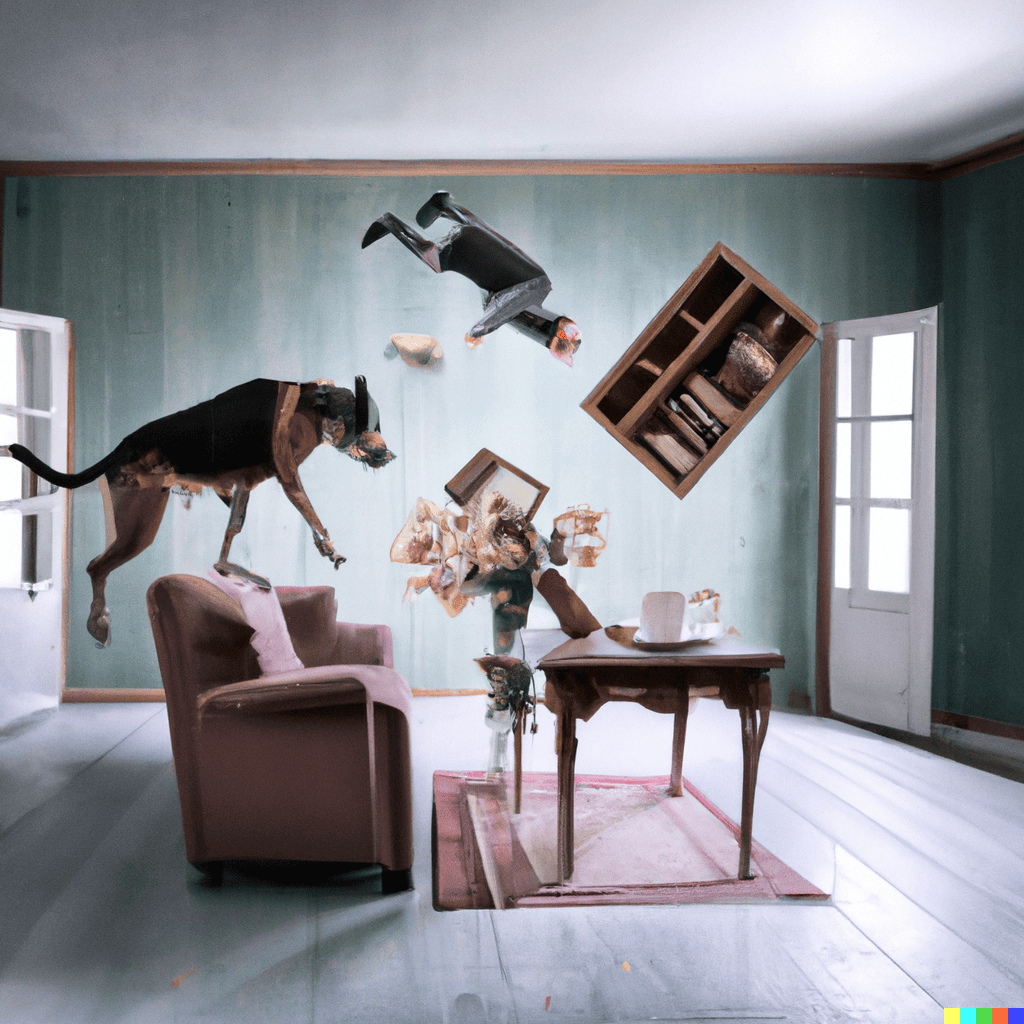}
    \caption{an image by DALL-E 2}\label{fig:d2}
    \label{subfig:dalle}
  \end{subfigure}
  \begin{subfigure}[!ht]{0.35\textwidth}
    \centering
    \includegraphics[width=\linewidth]{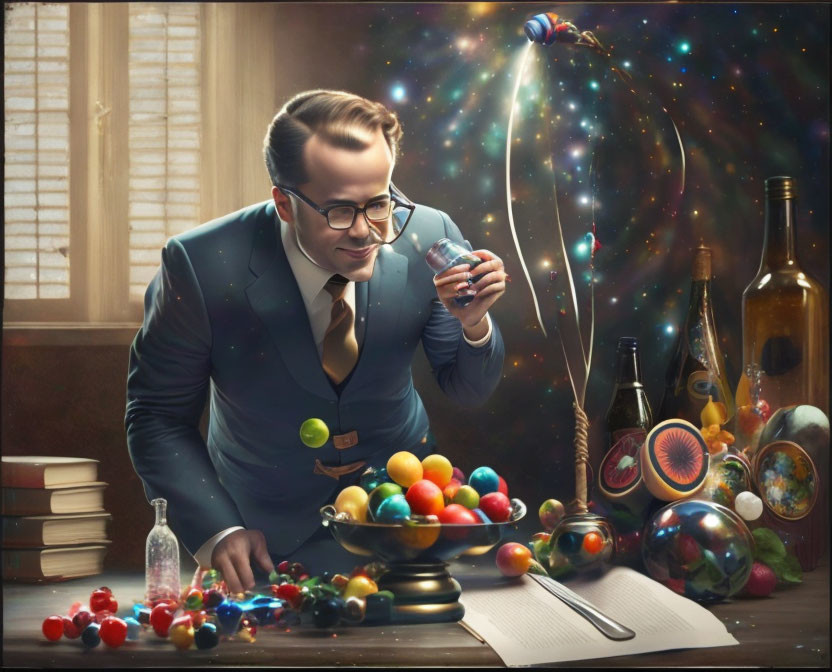}
    \caption{an image by Deep Dream Generator}\label{fig:dd}
    \label{subfig:deepdream}
  \end{subfigure}
  \begin{subfigure}[!ht]{0.29\textwidth}
    \centering
    \includegraphics[width=\linewidth]{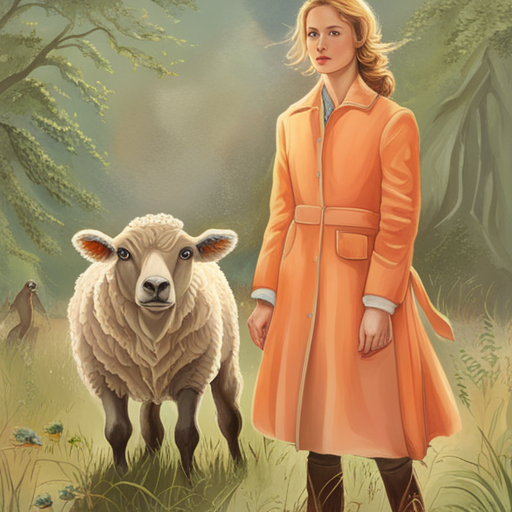}
    \caption{an image by DreamStudio}\label{fig:ds}
    \label{subfig:DreamStudio}
  \end{subfigure}
  \caption{Examples for the comparison of aesthetics of generated images}
  \label{fig:image_comparison}
\end{figure}
\subsubsection{Generated Images}
 As the prompt size grows, Dall E generates more surrealistic-like images with good color combinations. The generated images exhibit a more frequent interplay of light and dark shades, which makes it easy to raise attention. We also noticed that the model grasps concepts of the input prompts well, though not always the provided numbers. DALL-E's image generation is somewhat limited due to its production of squared images. Some generated images are particularly interesting. For example, Figure \ref{fig:d2} exhibits the idea of Dali's loss of gravity.

As for the second model, the Deep Dream Generator, we observe that it produces a wide range of colors in its generated images, e.g. Figure \ref{fig:dd}. Deep Dream often shows the capacity to incorporate colors from a base image, resulting in outcomes with similar color combinations, likely due to the influence of the input base image (e.g. Figure \ref{subfig:deepdream}). It showcases a variety of unexpected object combinations and surrealistic-like elements, even when provided with simple word prompts through YOLO labeling.

DreamStudio, the final model in our analysis, exhibits certain limitations in its capabilities. While some of the generated images are surrealistic-like, there are instances where the images lack any surrealistic-like elements (e.g. Figure \ref{fig:ds}). Despite this, DreamStudio excels at capturing the color, style, and positioning of objects from the base image when generating new images.

\subsubsection{Generated Prompts}
Our experiments show that ChatGPT-generated prompts consistently feature YOLO labelings, as previously mentioned. The resulting prompts are remarkably clear and easily comprehensible, demonstrating the ability to produce surrealistic-like content implicitly, even without explicitly using the term ``surrealistic.'' Notably, the prompts exhibited uniqueness and creativity, encompassing various words, actions, objects, and individuals. However, some generated prompts could be more creative.

\section{Discussion}
\label{sec:Discussion}
The statistical analysis and observations consistently support that  DALL-E excels in surrealistic-like image generation, when provided with a 50-word prompt generated by ChatGPT in comparison to other settings. DALL-E's image generation capability gets better when longer prompts are provided since the model relies on textual input to produce images. The results indicate that making use of  generative language models for surrealistic-like prompts positively impacts the performance of the generative image model. The prompts generated by ChatGPT receive a high number of votes from the participants. This suggests that ChatGPT's natural language processing capabilities contribute to producing prompts that was considered surrealstic-like by our panel of artists, resulting in visually captivating and surrealistic-like images.

Furthermore, our analysis indicates a correlation between prompt length and participants' preferences. Longer prompts generated by ChatGPT received more votes, indicating that comprehensive and detailed textual guidance plays an important role in achieving superior surrealistic-like image outcomes. Longer prompts offer richer contextual information, enabling generative image models to produce more surrealistic-like results. The integration of language models in guiding image generation has the potential to result in even more sophisticated surrealistic-like artworks, inspiring artists and creators to explore novel artistic horizons. We observe that including the artist's name as a prompt did not significantly impact image generation for the chosen models. However, it is valuable that DALL-E, being a text-based generative model, exhibited a unique advantage in capturing the essence of the base image's artist. Since DALL-E does not rely on input images, incorporating the artist's name allowed it to interpret and incorporate the stylistic elements associated with the mentioned artist. As a result, DALL-E generated images with a higher resemblance to the specified artist's style, showcasing its adaptability in producing artistically influenced outputs based solely on textual descriptions. Despite not having a substantial impact on other models, this capability of DALL-E highlights its potential to leverage artistic references effectively in the image generation process.

The preference for downscaled base images as input to the model increased in votes across generations, when contrasted with the utilization of blurred base images as input. This indicates that the downsizing technique may outperform blurring in enhancing the quality of generated surrealistic artworks. By downsizing the base images to one-fourth of their original size and using a 50-word long prompt, we introduce a novel element that resonates with participants.  It's also critical to recognize possible bias created by both the choice of base images and their application in our study. The selection of the base images carries a subjectivity that may unintentionally impact the results that are produced. Due to the intrinsic artistic styles embodied in these base images, the generative process may unintentionally lean towards those particular aesthetics. Furthermore, the use of terminology like "surrealism" while creating prompts carries the risk of prejudice. The use of this phrase might generate a bias towards adhering to accepted ideas about surrealistic art, which would then direct the model's interpretation and creative output.

Our analysis shows that all three models can generate some images that are appealing to artists. This indicates that they can use such language-vision models to boost their creativity. The combination of human and AI creativity can spark new ideas and inspire artworks. Knowing which model and settings to use can save time for artists. 

The outcomes from the particular sub-questions may not provide straightforward conclusions due to the exclusive focus on DreamStudio and Deep Dream Generator in Survey Part 3. Nevertheless, we can discern a notable trend: Deep Dream Generator garnered the highest number of votes in response to Part C questions, despite the constraint of allowing only a single choice. This contrast with Sub-questions A and B, which offered multiple selections, suggests that the clarity and simplicity of the Part C inquiries likely facilitated participants' choices. It can be inferred that the nature of the question, being comprehensible and straightforward with a single-response format, contributed to this outcome. Despite Deep Dream Generator receiving a significant number of votes in the sub-questions, DALL-E remains the premier choice as a surrealistic image generator. This distinction arises from DALL-E's exclusion in the final segment of the survey, where it particularly shines in specific configurations.

\section{Conclusion and Future Work}
\label{sec:Conclusion}
In conclusion, our study evaluates the capabilities of three different generative models. We explore various settings for each model to determine the most suitable configurations, and to test each model effectively for optimal image generation. Addressing our first research question \textbf{SRQ1} (Which image generative models are best suited for the generation of surrealistic-like images), we present the participants' preferences when shown the images. The majority of participants selected DALL-E with the 50-word prompt setting(PS4) for producing the best surrealistic-like images. \textbf{This finding highlights DALL-E as the most proficient model among the three for generating surrealistic-like imagery, particularly when guided by ChatGPT-generated prompts.} Furthermore, when images are generated solely by DALL-E without input from ChatGPT, the resulting images tend to be less  surrealistic-like. On the other hand, Deep Dream Generator, by default, tends to produce surrealistic-like images even without the need for external input or creative prompts solely from YOLO. This behavior is intrinsic to the Deep Dream Generator, which enhances patterns and details in an image in a way that often leads to surrealistic-like images.

For our second \textbf{SRQ2} (Can generative language models make meaningful suggestions by providing prompts), the results obtained indicate that the ChatGPT-generated prompts are highly preferred among participants compared to YOLO labelings alone. The preference for ChatGPT-generated prompts surpassed that of YOLO labelings, suggesting that generative language models can provide meaningful suggestions for image generation. 

As for the final sub-research question \textbf{SRQ3} (How can we improve the generation of images), is addressed based on our results, revealing that settings involving a ChatGPT prompt of 50 words and a downscaled base image did not receive a high preference. However,the downscaled version is preferred over the blurred version. Although downscaled base images show some improvement, they do not have a significant impact on DreamStudio and Deep Dream Generator. Additionally, to reaffirm our finding that PS4 with DALL-E is the best configuration, we decided to further explore DALL-E's capabilities by generating additional surrealistic-like images. For this purpose, we enriched the prompts with names of renowned surrealistic artists. Now, we present these new surrealistic-like creations, achieved by DALL-E with the 50-word prompt setting combined with the influence of surrealistic artists.

\begin{figure}
    \centering
    \begin{subfigure}{0.25\textwidth}
        \includegraphics[width=\linewidth]{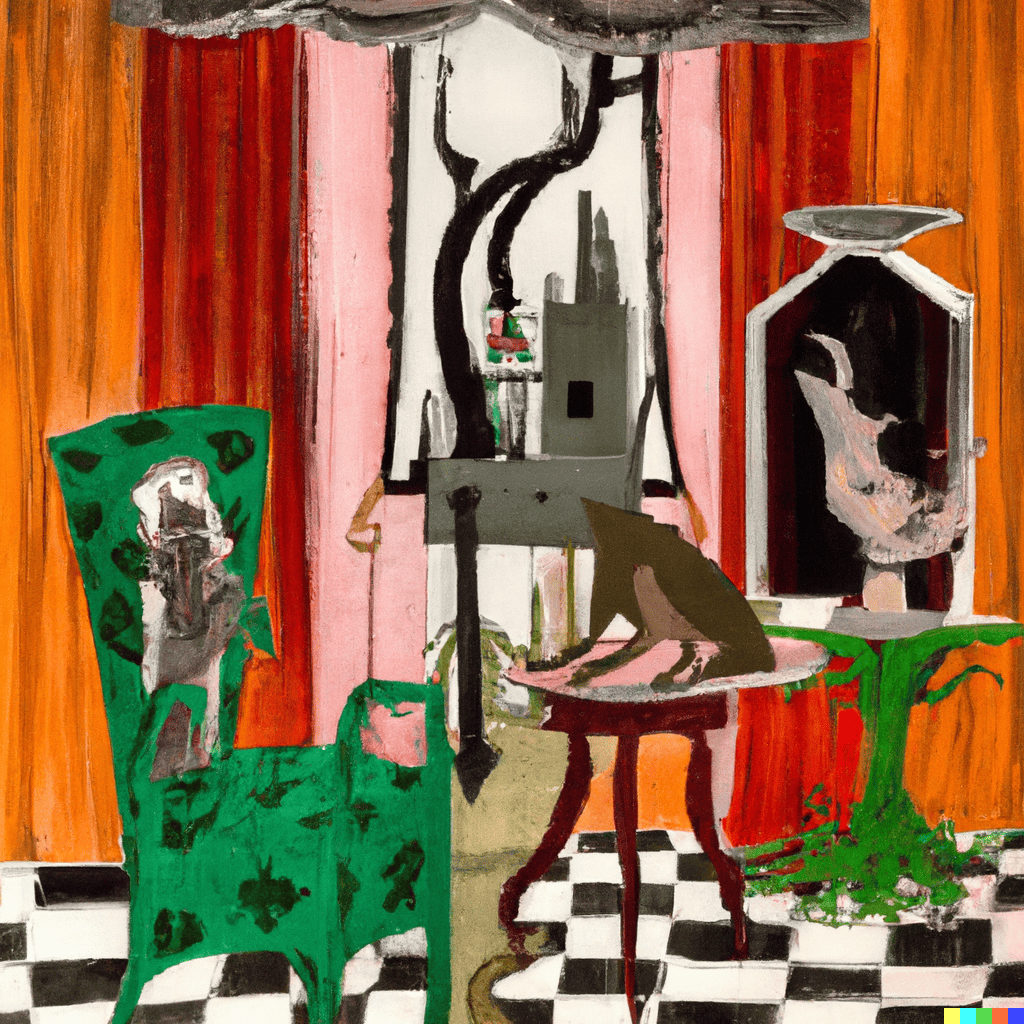}
        \caption{An example using the name Max Ernst}
        \label{fig:image1}
    \end{subfigure}
    \hfill
    \begin{subfigure}{0.25\textwidth}
        \includegraphics[width=\linewidth]{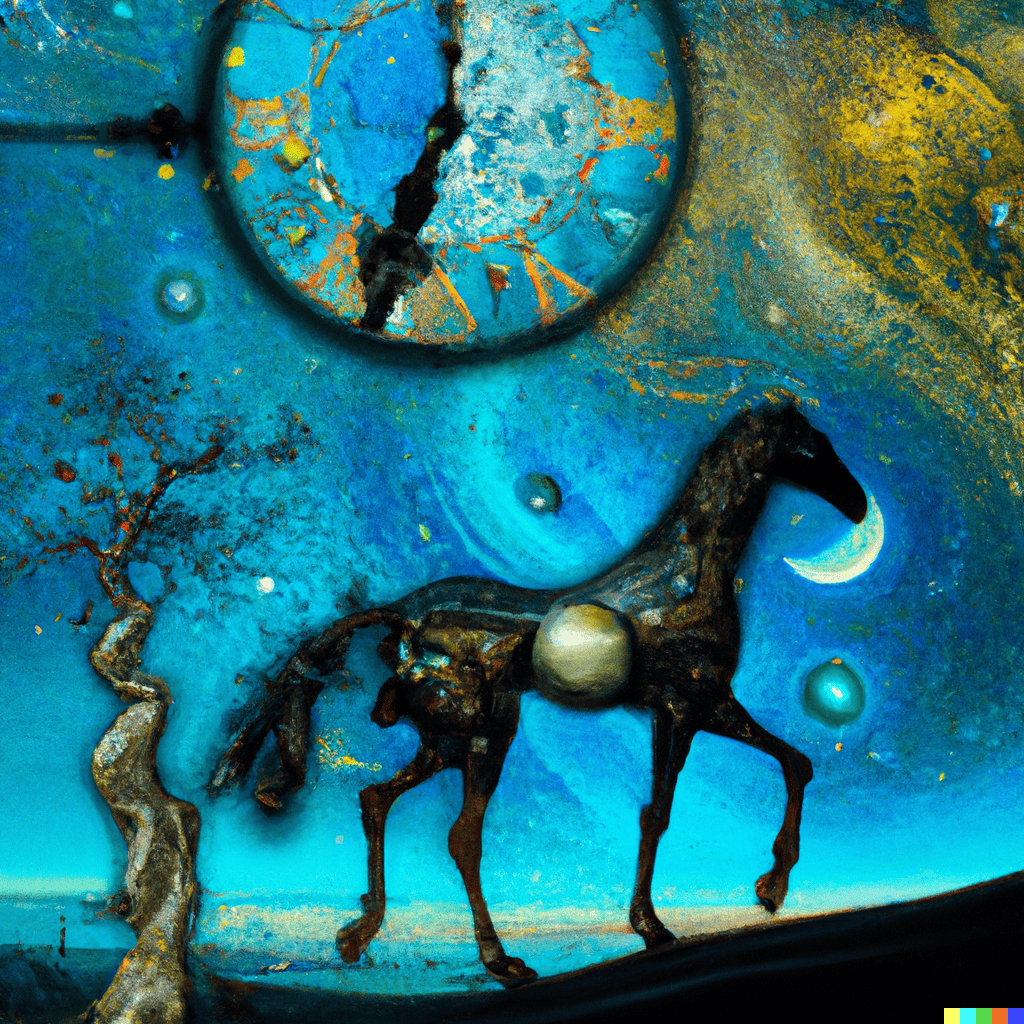}
        \caption{An example using the name Salvador Dalí}
        \label{fig:image2}
    \end{subfigure}
    \hfill
    \begin{subfigure}{0.25\textwidth}
        \includegraphics[width=\linewidth]{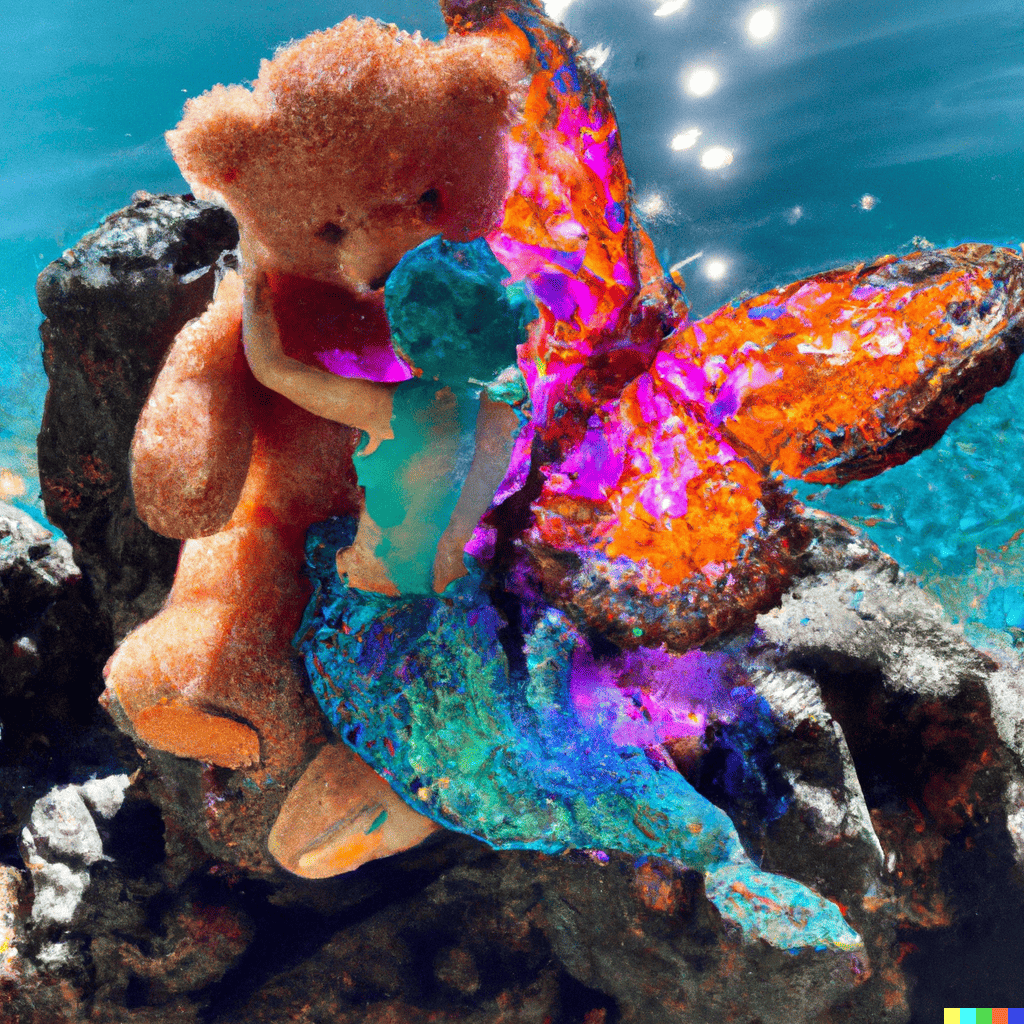}
        \caption{An example using  the name Juan Miro}
        \label{fig:image3}
    \end{subfigure}
    \caption{Images generated by Dall E 2 with the prompt of 50 words by ChatGPT including artists' names (see the supplementary material for the prompts)}
    \label{fig:surreal_images}
\end{figure}

In Figure \ref{fig:surreal_images}, we observe that incorporating surrealistic artist names along with the ChatGPT generated prompt significantly enhances the surrealistic quality of the images. The presence of surrealistic artist names seems to stimulate the generation process, inspiring the model to produce artwork with a heightened level of surrealistic-like characteristics. The combination of the model's creative capabilities and the influence of renowned surreal artists creates better surrealistic-like images.

By combining their strengths, new generative models can be explored for use in the future work, which will lead to the exploration of more hybrid methodologies. Giving ChatGPT the assignment of creating artworks that reflect the distinctiveness of particular artists with their favorite color schemes and style components, for instance, will give the generative process more depth. The generations could become more surrealistic-like by experimenting with the already-present surrealistic-like components from the paintings. For instance, melting things, varied perspectives, and unusual scales. Even while some may believe these tools can be limited and can constrain artists, there is great potential for unique styles and novel forms of what we call surrealism. In the end, the development of surrealistic-like images can be enhanced and go further with the help of creative AI. This symbiotic relationship may allow artists to explore innovative art forms that go beyond traditional boundaries or find their distinct style in AI-generated art.  In the latest version of ChatGPT-4o, a user can generate images with prompt with a specified style (e.g. Salvador Dalí). However, this is not the case for all artists. It is unable to generate an image in the style of Picasso due to content policy restrictions (tested on 18th Dec 2024). This new feature could be explored and compared against our results.

\bibliography{bib}

\end{document}